\documentclass[sigconf]{acmart}
\usepackage{booktabs} 
\usepackage{siunitx} 
\usepackage{indentfirst}
\usepackage{hhline}

\usepackage{lipsum}
\usepackage{graphicx}
\usepackage{footnote}

\usepackage{amssymb}

\usepackage{footnote}
\usepackage{amsmath}
\usepackage{mathtools}
\usepackage{mathrsfs}                                      
\usepackage{siunitx} 
\usepackage{comment}
\usepackage{wrapfig} 
\usepackage[subrefformat=parens,labelformat=parens]{subfig}
\usepackage{bm,bbm}
\usepackage{multirow}
\usepackage{makecell}
\usepackage{threeparttable,booktabs}
\usepackage{blkarray}
\usepackage{tikz}
\usetikzlibrary{positioning,calc,fit,decorations.pathmorphing,shapes.geometric,shapes.gates.logic.US,calc}
\usepackage{tikz-3dplot}
\usepackage{balance}
\usepackage{courier}                                       
\usepackage{cleveref}                                      
\usepackage[mathcal]{eucal}
\usepackage[]{algpseudocode}                               
\algrenewcommand\textproc{\texttt}
\makeatletter\let\float@addtolists\relax\makeatother
\usepackage{algorithm}
\algblockdefx[pardo]{pardo}{endpardo}%
   [1]{\textbf{For} #1 \textbf{do in parallel}}%
   {\textbf{Endfor}}
\usepackage{filecontents}                                  
\usepackage{pgfplots}
\usepackage{pgfplotstable}
\usepackage{pgf-pie}
\usepgfplotslibrary{groupplots}
\pgfplotsset{compat=newest}
\usepackage[figuresright]{rotating}
\usepackage{xcolor,colortbl}                               
\usepackage[inline]{enumitem}                              
\setlist{leftmargin=5.08mm}

\renewcommand{\vec}[1]{\boldsymbol{#1}}

\theoremstyle{plain}

\theoremstyle{definition}

\newtheorem{myproblem}{\textbf{Problem}}

\algrenewcommand\textproc{\texttt}

\usepackage[skip=1pt]{caption}            
\definecolor{CUHKorange}{RGB}{244,106,18} 
\definecolor{CUHKblue}{RGB}{0,111,190}    
\definecolor{CUHKgreen}{RGB}{0,127,128}   
\definecolor{CUHKred}{RGB}{228,46,36}     
\definecolor{CUHKyellow}{RGB}{198,148,34} 
\definecolor{CUHKdark}{RGB}{114,44,114}   
\definecolor{CUHKmiddle}{RGB}{144,44,144} 

\usepackage{tcolorbox}
\tcbuselibrary{skins,breakable}
    {\endtcolorbox}

\graphicspath{{./figs/}}

\copyrightyear{2024}
\acmYear{2024}
\setcopyright{acmlicensed}\acmConference[ICCAD '24]{IEEE/ACM International Conference on Computer-Aided Design}{October 27--31, 2024}{New York, NY, USA}
\acmBooktitle{IEEE/ACM International Conference on Computer-Aided Design (ICCAD '24), October 27--31, 2024, New York, NY, USA}
\acmDOI{10.1145/3676536.3676752}
\acmISBN{979-8-4007-1077-3/24/10}

\begin{document}
\date{}

\title{
    SEM-CLIP: Precise Few-Shot Learning for Nanoscale Defect Detection in Scanning Electron Microscope Image
}


\author{Qian Jin}
\affiliation{%
  \institution{Zhejiang University}
  \city{Hangzhou}
  \country{China}}

\author{Yuqi Jiang}
\affiliation{%
  \institution{Zhejiang University}
  \city{Hangzhou}
  \country{China}}

\author{Xudong Lu}
\affiliation{%
  \institution{Zhejiang University}
  \city{Hangzhou}
  \country{China}}

\author{Yumeng Liu}
\affiliation{%
  \institution{Zhejiang University}
  \city{Hangzhou}
  \country{China}}

\author{Yining Chen}
\affiliation{%
  \institution{Zhejiang University, HIC-ZJU}
  \city{Hangzhou}
  \country{China}}

\author{Dawei Gao}
\affiliation{%
  \institution{Zhejiang University, HIC-ZJU}
  \city{Hangzhou}
  \country{China}}

\author{Qi Sun$^{\#}$}
\affiliation{%
  \institution{Zhejiang University}
  \city{Hangzhou}
  \country{China}}
\email{qisunchn@zju.edu.cn}

\author{Cheng Zhuo$^{\#}$}
\affiliation{%
  \institution{Zhejiang University}
  \city{Hangzhou}
  \country{China}}
\email{czhuo@zju.edu.cn}

\thanks{$^\#$ Corresponding authors.}

\begin{abstract}

In the field of integrated circuit manufacturing, the detection and classification of nanoscale wafer defects are critical for subsequent root cause analysis and yield enhancement. The complex background patterns observed in scanning electron microscope (SEM) images and the diverse textures of the defects pose significant challenges. Traditional methods usually suffer from insufficient data, labels, and poor transferability. In this paper, we propose a novel few-shot learning approach, SEM-CLIP, for accurate defect classification and segmentation. SEM-CLIP customizes the Contrastive Language-Image Pretraining (CLIP) model to better focus on defect areas and minimize background distractions, thereby enhancing segmentation accuracy. We employ text prompts enriched with domain knowledge as prior information to assist in precise analysis. Additionally, our approach incorporates feature engineering with textual guidance to categorize defects more effectively. SEM-CLIP requires little annotated data, substantially reducing labor demands in the semiconductor industry. Extensive experimental validation demonstrates that our model achieves impressive classification and segmentation results under few-shot learning scenarios.

\end{abstract}


\maketitle
\pagestyle{plain}

\section{Introduction}
\label{sec:introduction}

Semiconductor manufacturing is a complex and multifaceted process where defects occur due to ill processes or equipment issues. To provide real-time monitoring for the fabrication, SEM images are captured and then classified based on the appearance of the defects, helping the defect detection and root cause analysis. 
Unlike rough wafer-level defect maps, SEM images can provide more detailed characteristics of defects, thereby helping to determine the specific process steps and equipment. Currently, defect detection primarily relies on manual efforts, making it both cumbersome and error-prone. Developing an automated defect detection system has become a trend.

The current wafer surface defect detection and classification research predominantly employs supervised learning methods, requiring substantial amounts of data and detailed annotated labels. Some methods are presented to classify defects \cite{chen2008defect, chang2013hybrid, cheon2019convolutional}. Furthermore, some segmentation methods are proposed to provide detailed location and shape information \cite{ronneberger2015u, badrinarayanan2017segnet, nag2022wafersegclassnet}. Although these methods achieve outstanding performance, they usually require many annotated data for training, resulting in heavy workloads. Besides, these methods also suffer from poor transferability for new defect detection due to a lack of adequate training data. Annotated data is always precious in industry.

\begin{figure}[tb!]
  \centering
  \includegraphics[width=\linewidth]{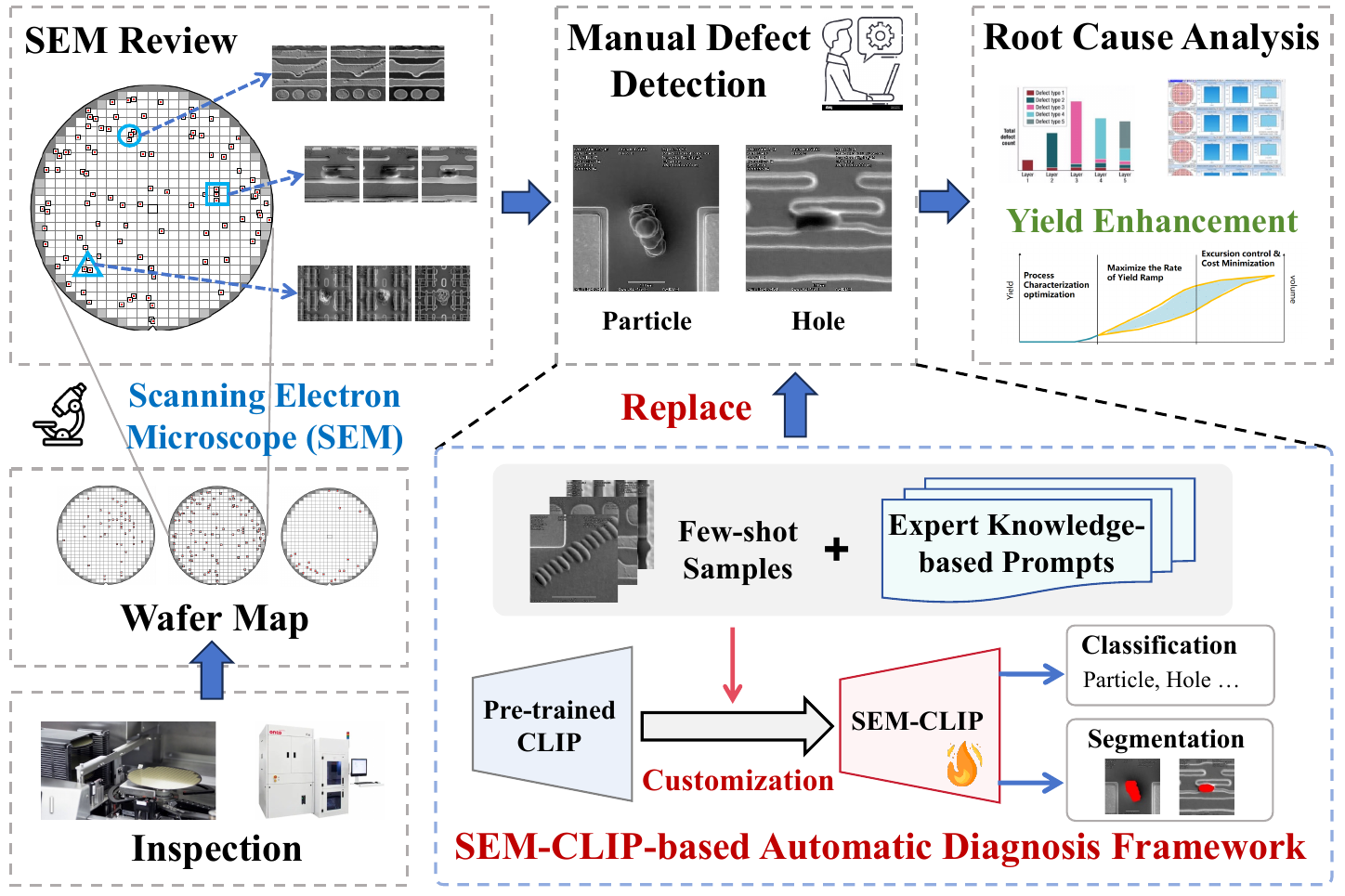}
  \caption{The workflow of SEM image defect analysis. We replace the cumbersome manual defect detection flow with our automatic SEM-CLIP method, substantially enhancing defect detection performance with few-shot learning as the shining point. 
}
  \label{fig:workflow}
\end{figure}

Consequently, there has been a shift in the field of industrial defect detection toward unsupervised or self-supervised anomaly segmentation methods \cite{zavrtanik2021draem, jiang2022masked, jiang2022softpatch, wang2023multimodal}. These approaches only require normal samples to learn their distribution, and they detect anomalies by calculating the distributional differences between test samples and normal samples. However, this method still requires a substantial number of normal samples for training. Due to the highly variable backgrounds where defects occur, there are significant differences among normal samples, making applying this approach in wafer surface defect detection scenarios challenging.

Recently, pre-trained vision-language models like CLIP \cite{radford2021learning} and SAM \cite{kirillov2023segany} have rapidly advanced, utilizing prompts to access stored prior knowledge and thus exhibiting strong zero-shot visual perception capabilities \cite{Li_Li_Li_Niebles_Hoi_2022}. Considering this, we are exploring using a CLIP model-based approach to address data scarcity issues. However, given the unique aspects of integrated circuit application scenarios, the text-image pairs used in network pre-training may contain minimal or no SEM images of semiconductors. Consequently, it becomes essential to adjust the base structure of the CLIP model and to incorporate a small number of SEM images of both normal and anomalous samples as support images for the target categories. These adaptations will enable the model to more effectively recognize and classify the specific types of defects encountered in semiconductor manufacturing.

This strategy allows us to leverage the model's inherent ability to understand complex visual concepts through minimal samples, adapting it to the specific requirements of semiconductor manufacturing. We can create a more efficient and effective model for detecting and classifying wafer surface defects without heavily relying on large, annotated datasets. To this end, we propose SEM-CLIP, a crafted CLIP method for defect detection, following the few-shot learning mechanism. The contributions of our work are summarized as follows:
\begin{itemize}
    \item We propose a novel few-shot learning-based approach, SEM-CLIP, for accurate SEM image defect classification and segmentation with little data and label requirements. To the best of our knowledge, it is the first few-shot learning work for SEM-level IC defect detection tasks. 
    \item We customize the Contrastive Language-Image Pretraining model to focus on the defect areas and adopt a novel feature extraction method by adding $V$-$V$ attention blocks to minimize the complex background distractions and improve the segmentation accuracies.
    \item Prompts enriched with expert knowledge are crafted and employed as prior information to guide both classification and segmentation processes. Feature engineering with textual guidance is incorporated with a classification head to boost the classification performance. 
    \item We conduct comprehensive experiments across various few-shot settings, benchmarked on an in-house SEM image defect dataset. The results demonstrate that our method significantly outperforms others in terms of iAUROC, pAUROC, and $F1$-$max$ scores. For instance, 
    SEM-CLIP surpasses the recent SOTA method PromptAD, showing improvements of $2.0\%$, $1.3\%$, and $21.1\%$, respectively, under the 10-shot setting. Our approach will help fabs alleviate the issues of insufficient labeling and expensive labor, thereby facilitating intelligent manufacturing.
    
\end{itemize}

\section{Preliminaries}
\label{sec:preliminaries}

\subsection{Pre-trained Vision-language Model}

Vision-language models process and integrate visual and textual data, enabling tasks that require a cohesive understanding of both domains. The CLIP model \cite{radford2021learning}, which was pre-trained on 400 million image-text pairs, has robust generalization and enables it to utilize natural language to refer to learned visual concepts. These Transformer-based encoders \cite{DNN-2017NeurIPS-Attention} project features into a shared embedding space where similarity is computed, guided by a contrastive loss function that aligns matching pairs and separates non-matching pairs. This design allows CLIP to generalize effectively across various tasks without task-specific training, demonstrating its flexibility in downstream applications \cite{luddecke2022image, zhou2022maskclip, Zhou_Zhang_Lei_Liu_Liu_2022, Chen_Si_Zhang_Wang_Wang_Tan_2023}.


\subsection{Wafer Surface Defect Detection}


Defect detection is essential for improving yields in integrated circuit fabrication. Traditional research has focused on wafer maps, where faulty chips are marked with colors based on test results. While these maps can provide spatial insights into defects, the increasing complexity of chip components has made wafer map-level detection more challenging and less precise \cite{Wafer-2020TSM-DCNN, Wafer-2022TSM-FusionTransformer, Wafer-2023ASPDAC-Survey, ma2023review}.
To address these limitations, magnified imaging techniques like scanning electron microscopy (SEM) are crucial for closely examining wafer surfaces. As shown in \Cref{fig:workflow}, advanced methods are needed to accurately detect, classify, and analyze microscopic defects, pinpointing the exact process steps where defects originate.

\subsection{SEM Image Defect Data}

\begin{figure}[h]
  \centering
  \includegraphics[width=0.98\linewidth, trim={5cm 4.8cm 5cm 3.1cm}, clip]{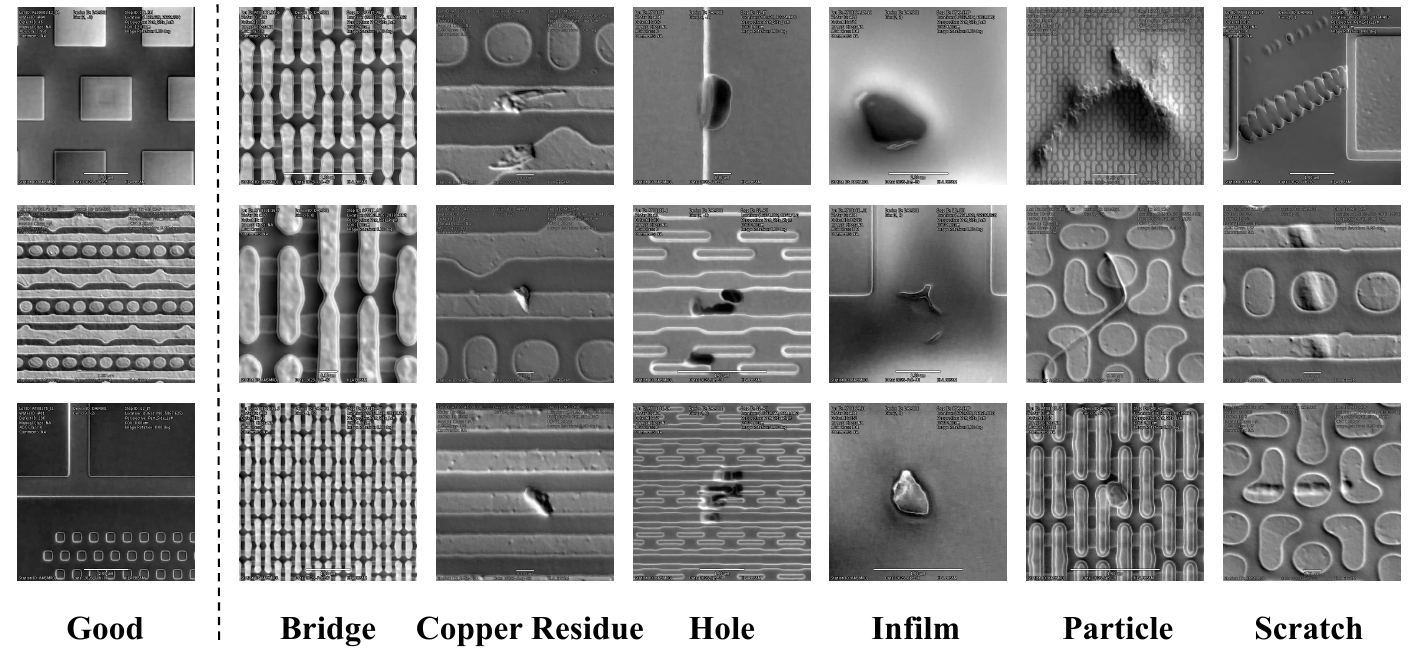}
  \caption{Non-defect and defective images.} 
  \label{dataset}
\end{figure}

In the absence of a public SEM Image dataset, we collect some data from an in-house 12-inch, 55$nm$ CMOS fabrication line.
The dataset includes 1332 grayscale images, with 226 non-defective and 1106 defective images, categorized into six common defect types: 59 bridges, 141 copper residues, 230 holes, 77 infilm defects, 455 particles, and 144 scratches. \Cref{dataset} illustrates some examples.

\subsection{Related Work}

Wafer surface defect detection was traditionally performed by engineers, relying on expertise that is time-consuming and inconsistent. With advancements in artificial intelligence, deep learning techniques have become highly effective for this task \cite{gao2022review}. Several classification approaches have been developed. Chen \textit{et al.} proposed a defect recognition algorithm using PCA and SVM \cite{chen2008defect}. Chang \textit{et al.} utilized SVM with features like smoothness and texture \cite{chang2013hybrid}. Cheon \textit{et al.} introduced a CNN model for feature extraction \cite{cheon2019convolutional}. Defect segmentation is crucial for determining defect locations and sizes. Encoder-decoder networks like UNet \cite{ronneberger2015u} and SegNet \cite{badrinarayanan2017segnet} are commonly used. Han Hui \textit{et al.} combined a Region Proposal Network (RPN) with UNet for defect area suggestion \cite{han2020polycrystalline}. Subhrajit Nag \textit{et al.} introduced WaferSegClassNet, which performs both classification and segmentation \cite{nag2022wafersegclassnet}. Recently, Vic De Ridder \textit{et al.} applied diffusion models to predict and reconstruct masks for semiconductor defects, achieving high precision but at a high computational cost, and with limitations in handling only a single defect type \cite{de2023semi}.

Despite these advancements, these methods rely heavily on large amounts of accurately labeled data, which is scarce, and they struggle with transferring to new defect types.

\subsection{Few-shot Anomaly Detection}

Traditional anomaly detection relies on extensive training data, which limits its effectiveness in dynamic environments with diverse anomaly types. Recent research has focused on using few or zero samples to overcome these challenges.
Ding \textit{et al.} introduced DRA \cite{ding2022catching}, which, although not specifically mentioning the concept of few-shot learning, effectively identifies both seen and unseen anomalies through disentangled representations by learning from a small number of labeled samples. 
Recent studies show that pre-trained vision-language models such as CLIP can significantly enhance performance in this task.
Jeong \textit{et al.} developed WinCLIP \cite{Jeong_2023_CVPR}, the first framework to use visual language models for few-shot anomaly detection, integrating state words and prompt templates with a novel window-based technique for improved performance. Gu \textit{et al.} introduced AnomalyGPT \cite{gu2024anomalygpt}, leveraging large vision-language models trained on simulated anomalies to effectively locate them.
Chen \textit{et al.} proposed CLIP-AD (zero-shot) \cite{chen2023clip}, and Li \textit{et al.} introduced PromptAD (few-shot) \cite{li2024promptad}, both using dual-path models and feature surgery to enhance CLIP’s anomaly detection capabilities.

These studies push the boundaries of traditional anomaly detection, showing how few-shot learning can rapidly and effectively address dynamic, data-scarce environments. Our research extends the CLIP method to support SEM image defect detection. 


\subsection{Problem Definition}
\begin{myproblem} [Few-shot Learning for SEM Image Defect Detection]

Given dataset of $N$-way $K$-shot SEM images $\vec{X} = \{\vec{x}_1, \vec{x}_2 \cdots, \vec{x}_{K \cdot N}\}$, annotated with classification labels $ \vec{Y}^c = \{\vec{y}_1^c, \vec{y}_2^c, \cdots, \vec{y}_{K \cdot N}^c\}$ and segmentation masks $\vec{Y} ^s = \{\vec{y}_1^s, \vec{y}_2^s, \cdots, \vec{y}_{K \cdot N}^s\}$. Typically, $N$ represents the total number of categories in the dataset, including the ``good'' (non-defect) category, and all defect categories. $K$ is a small number denoting the number of images for each category, such as 1, 2, or 10, which is why this is referred to as few-shot learning. 
We aim to construct a model with few-shot learning capabilities based on the $\vec{X}$. It can generate accurate defect classification labels and pixel-level segmentation results for the $M$ SEM image testing set with $M \gg K$. By default, $N=7$ in our context without further explanations. 

\end{myproblem}

\section{SEM-CLIP Framework}
\label{sec:framework}

\begin{figure}[h]
  \centering
  \includegraphics[width=0.94\linewidth]{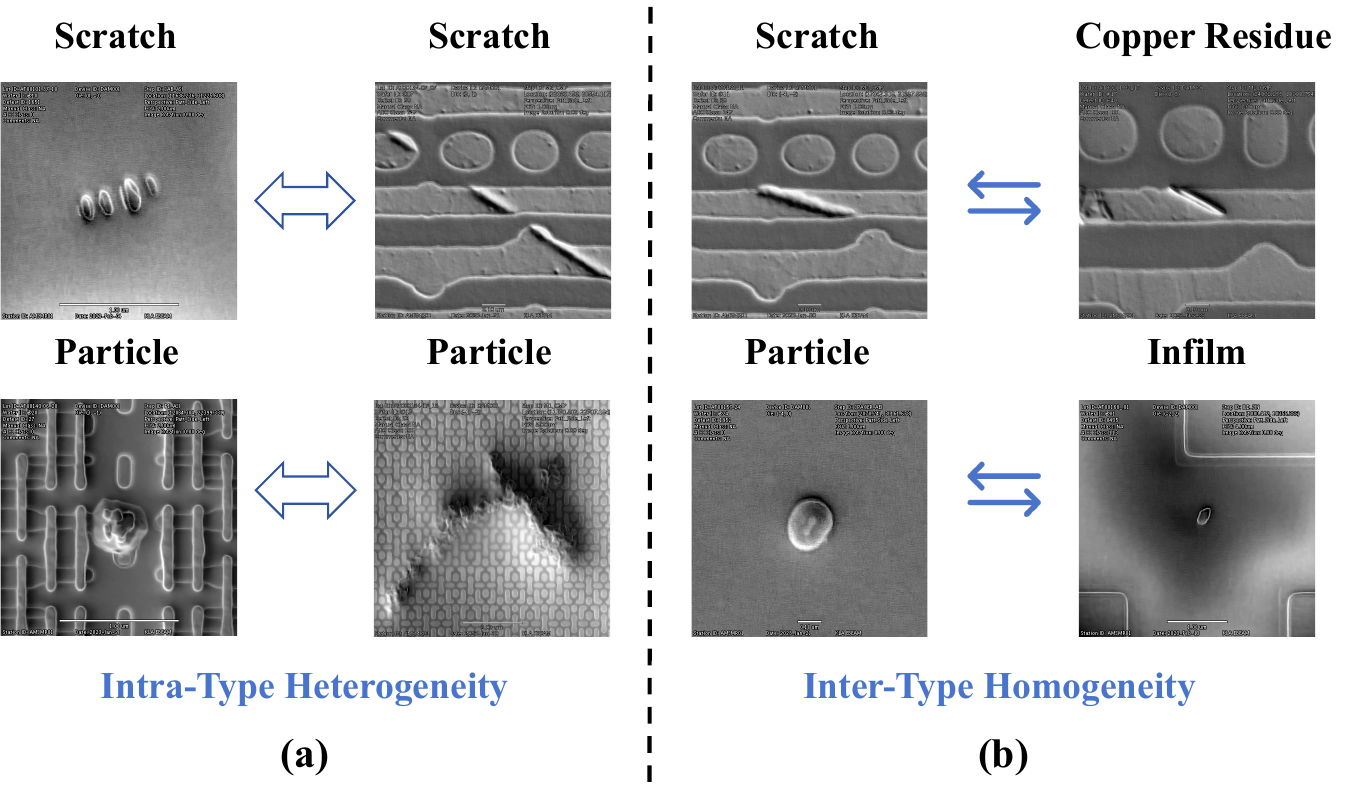}
  \caption{Complexity of defect morphologies. (a) Differences within the same type; (b) Similarity between different types. }
  \label{fig:defect-complexity}
\end{figure}

\begin{figure*}[tb!]
  \centering
  \includegraphics[width=0.86\textwidth]{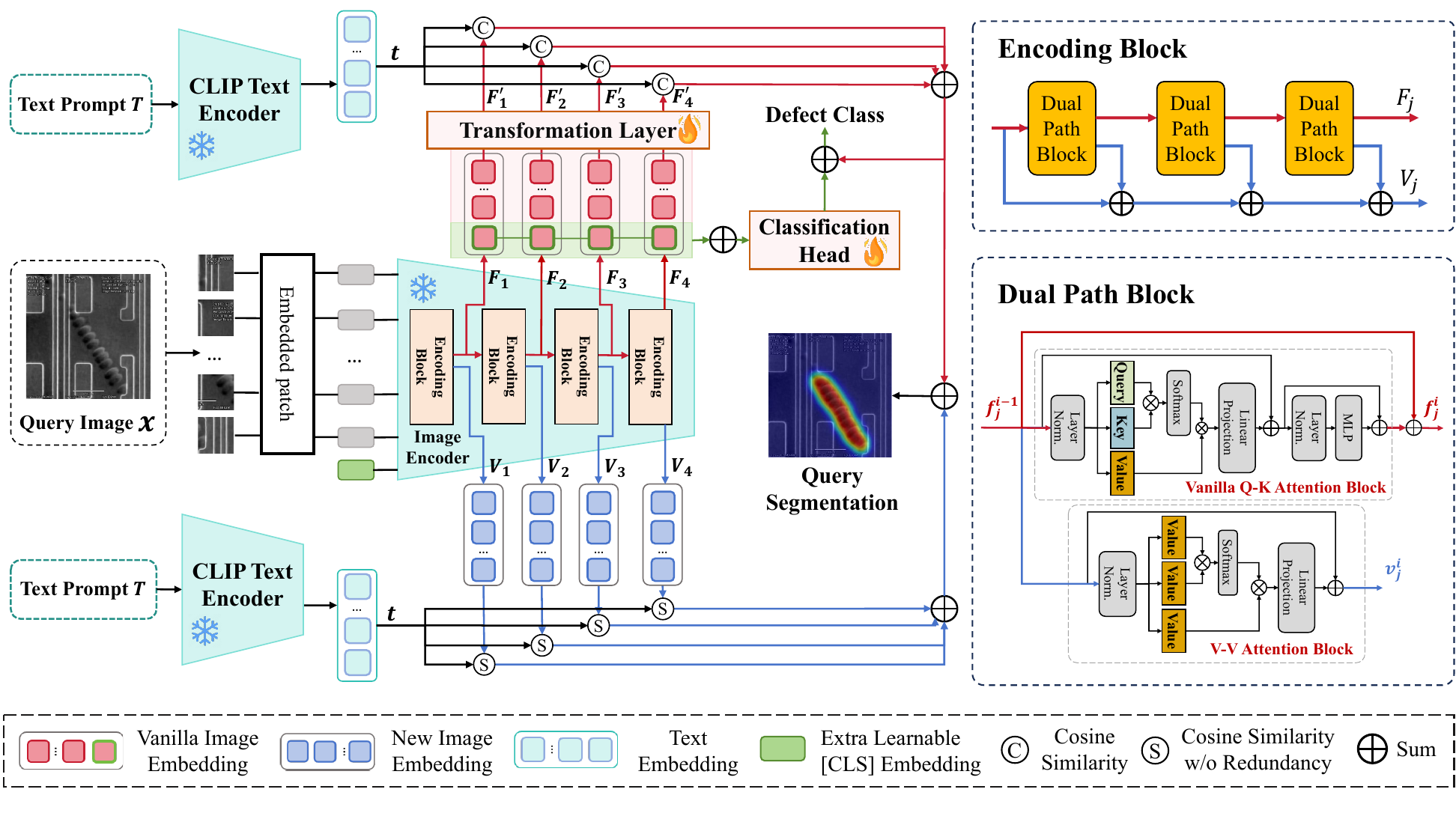}
  \caption{Our SEM-CLIP framework. }
  \label{fig:SEM-CLIP}
\end{figure*}

In this section, we introduce SEM-CLIP, as shown in \Cref{fig:SEM-CLIP}, specifically designed for classifying and segmenting wafer surface defects under the few-shot setting. Initially, we construct a text prompt incorporating expert knowledge regarding wafer surface defect patterns. This prompt enables us to avoid detailed labels for each sample. Following this, we implement a dual path block by adding a $V$-$V$ attention block to the transformer block within the vanilla ViT architecture \cite{dosovitskiy2020image}. 
We extract features at various levels from this architecture and employ a new method to remove redundant features to calculate similarity. Additionally, we fine-tune the Transformation Layer and Classification Head using few-shot samples, ultimately achieving precise defect classification and segmentation results.

\subsection{Text Prompt Design}

Due to the complexity of integrated circuit manufacturing processes, wafer surface defects can vary greatly in appearance, resulting in significant morphological differences within the same type of defect and similar textures between different types of defects \Cref{fig:defect-complexity}. Consequently, it is essential to utilize domain expert knowledge to refine the rough cues such as ``anomaly'' or ``defect'' into more detailed descriptions of defect morphologies by useful prior information about the target defect areas. 
For instance, defects of the ``scratch'' type typically appear as fine, long, linear marks in the \textit{back-end-of-line} (BEOL) processes but may manifest as fish-scale patterns in the \textit{front-end-of-line} (FEOL) processes. 
These elliptical depressions, which exhibit a continuous distribution, can easily be mistaken for hole-type defects without careful observation.

\begin{figure}[tb!]
  \centering
  \includegraphics[width=0.90\linewidth]{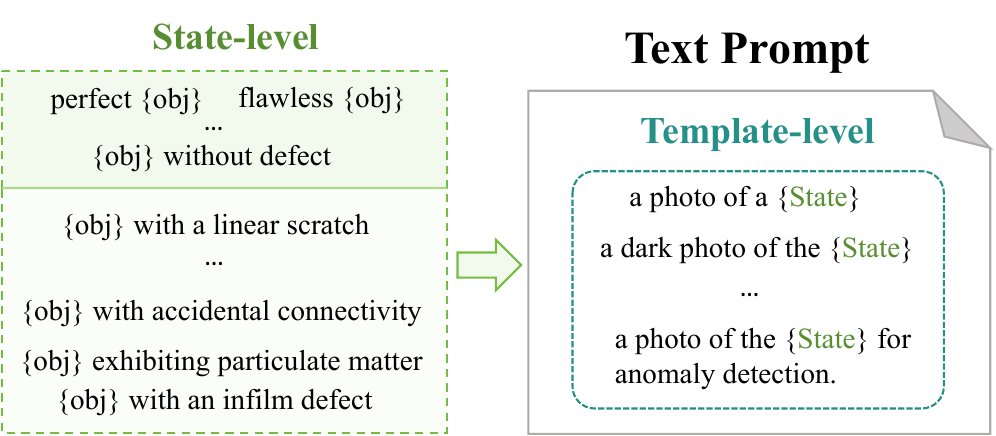}
  \caption{Text prompts are built on state-level prompts and template-level prompts. }
  \label{fig:text_prompt}
\end{figure}

This task employs a composite prompt structure, as illustrated in \Cref{fig:text_prompt}.  We decompose the prompts into template-level and state-level components, where the state-level prompts provide detailed descriptions of the possible appearances of each type of defect, such as ``\{ \} image with a linear scratch'' or ``\{ \} image with fish scale-shaped scratches''. Additionally, since scanning electron microscopes can produce blur due to focusing issues or variations in image brightness caused by different electron beam intensities, the template-level prompts can describe the effects on SEM images, such as ``a blurry photo of the \{ \}'' or ``a dark photo of a \{ \}''. Finally, by replacing the \textit{state} in the template-level prompts with the state-level prompts, we combine them to form the final text prompts.

The text prompts are designed and shared for all SEM images. During the practical application of our model and the analysis of query images, there is no need to adjust the prompts.

\subsection{Image Feature Extraction}

For SEM images, the variability and complexity of background patterns tend to interfere with defect detection, which is undesirable. Recent studies have reported that $Q$-$K$ self-attention \cite{DNN-2017NeurIPS-Attention} may lead to incorrectly establishing connections in semantically irrelevant areas
, resulting in dispersed attention \cite{li2023clip}. The vanilla self-attention mechanism is described as follows:
\begin{equation}
   \text{Attention}(\vec{Q}, \vec{K}, \vec{V}) = \text{Softmax}\left( \frac{\vec{Q}\vec{K}^\top}{\sqrt{d_k}} \right) \vec{V}.
\label{eq:QKV}
\end{equation}

In contrast, $V$-$V$ attention \cite{li2023clip}, by directly comparing and associating similar feature values, can more accurately focus on relevant feature areas, effectively reducing interference from the background. The $V$-$V$ attention is formulated as follows:
\begin{equation}
    \text{Attention}( \vec{V}, \vec{V}, \vec{V} ) = \text{Softmax}\left(\frac{\vec{V}\vec{V}^\top}{\sqrt{d_k}}\right) \vec{V}.
    \label{eq:VVV}
\end{equation}


Therefore, we modify the vanilla CLIP ViT \cite{dosovitskiy2020image} backbone for feature extraction by adding a branch while retaining the vanilla transformer structure. This branch incorporates the $V$-$V$ attention block, constructing a new dual-path block, and the encoding block is composed of $n$ dual-path blocks. The entire ViT backbone contains $m$ encoding blocks, as shown in \Cref{fig:SEM-CLIP}. 
Taking the $i$-th dual-path block within the $j$-th encoding block as an example, the input is $\vec{F_{j}^{i-1}}$, and it gives two outputs:
\begin{align}
\vec{F_{j}^{i}} &= \text{Arch}_{\vec{QKV}}(\vec{F_{j}^{i-1}}) + \vec{F_{j}^{i-1}}, \\
\vec{V_{j}^{i}} &= \text{Arch}_{\vec{VVV}}(\vec{F_{j}^{i-1}}),
\end{align}
where $\text{Arch}_{\vec{QKV}}$ and $\text{Arch}_{\vec{VVV}}$ signify the vanilla $QKV$ block and the $VVV$ block respectively. $\vec{F_{j}^{i}}$ and $\vec{V_{j}^{i}}$ denote the outputs of these two blocks.

The input of the $j$-th encoding block is the output of the last layer (the $n$-th dual-path block) of the $ \left(j-1\right) $-th encoding block:
\begin{align}
\vec{F_{j}^{0}} &= \vec{F_{j-1}^{n}} = \vec{F_{j-1}}.
\end{align}
Therefore, for the $j$-th encoding block, the output is:
\begin{align}
\vec{F_{j}} &= \vec{F_{j}^{n}} = \text{Arch}_{\vec{QKV}}(\vec{F_{j}^{n-1}}) + \vec{F_{j}^{n-1}}, \\
\vec{V_{j}} &= \sum_{i=0}^{n} \vec{V_{j}^{i}}.
\end{align}

We extract features at multi-levels from the output of the encoding block, resulting in $m$ vanilla image embeddings $[\vec{F_{1}}, \vec{F_{2}}, \cdots, \vec{F_{m}}]$ and $m$ new image embeddings $[\vec{V_{1}}, \vec{V_{2}}, \cdots, \vec{V_{m}}]$ transformed by $V$-$V$ attention.

Notably, the weights for our vanilla $QKV$ block are loaded from the weight file of the pre-trained CLIP image encoder. Additionally, the $VVV$ block parameters are directly copied from those in the $QKV$ block. We merely modify the method of data computation rather than the data itself. Therefore, retraining is unnecessary.

\subsection{Defect Segmentation}
\label{sub:Defect Segmentation}
When using a pre-trained CLIP model for zero-shot defect segmentation, the typical method is directly calculating the similarity between text and image embeddings to get a defect map. However, this approach is not suitable for our task. Although we have constructed a detailed textual prompt with expert knowledge,
the text still struggles to fully describe all information for corresponding images, especially for our unusual SEM images.
This means our problem cannot be addressed with a zero-shot approach. Instead, it requires few-shot samples for fine-tuning. In this study, we adopt a few-shot learning approach 
to improve the detection of SEM defects. The specific implementation details are as follows:


First, we utilize a pre-trained CLIP text encoder to transform the text prompt $\vec{T}$ into a text embedding $\vec{t}$:
\begin{align}
\vec{t} &= \text{TextEncoder}(\vec{T}).
\end{align}

As mentioned in the previous section, we modify the structure of the image encoder, resulting in two different types of image embeddings, denoted as $\vec{F}$ and $\vec{V}$. These embeddings are extracted from $m$ different levels.

\noindent
\textbf{Segmentation based on $\vec{F}$.}
The vanilla image embedding $ \vec{F} = \{\vec{f}^{CLS}, \vec{f}^1, \vec{f}^2, \ldots, \vec{f}^T \} $, where $ \vec{f}^{CLS}$ serves as the $CLS$ token aggregating the global features of the image, commonly used in image-level defect detection, consider applying it to defect classification tasks. 
$ \vec {F}[1:] = \{\vec{f}^1, \vec{f}^2, \ldots, \vec{f}^T\} $ contains more detailed information, so we use it for pixel-level defect segmentation. 

To enhance the model's understanding of our application scenario, we introduce a transformation layer fine-tuned with a few samples. Specifically, this transformation layer functions by mapping the image embeddings to a joint embedding space through a linear layer. The input for the mapping is represented as $[\vec{F_{1}}[1:], \vec{F_{2}}[1:], \cdots, \vec{F_{m}}[1:]]$, and the output is $[\vec{F_{1}^{'}}, \vec{F_{2}}^{'}, \cdots, \vec{F_{m}}^{'}]$. Taking the output image embedding $\vec{F_{j}}$ from the $j$-th encoding block as an example, the mapping process is as follows:
\begin{align}
\vec{F_{j}}^{'}=\text{Transformation}(\vec{F_{j}}[1:]).
\end{align}

For the transformed vanilla image embedding $\vec{F_{j}^{'}}$, we calculate its cosine similarity with the text embedding $\vec{t}$. The formula is as follows:
\begin{align}
\vec{s(F_{j}^{'}, t)} = \vec{\frac{F_{j}^{'} \cdot t}{\|\vec{F_{j}^{'}}\|_{2} \|t\|_{2}}},
\label{eq:similarity}
\end{align}
where $ \vec{F_{j}^{'} \cdot t} $ represents the dot product of $ \vec{F_{j}^{'}}$ and $ \vec{t}$, $ \vec{ \|F_{j}^{'}\|_{2} } $ and $\vec{ \|t\|_{2} }$ are the $L2$ norms of the vectors $\vec{F_{j}^{'}}$ and $ \vec{t}$ along $C$ dimension. 

After processing through the softmax layer, we obtain the defect map calculated from $\vec{F_{j}}$ of the $j$-th encoding block:
\begin{align}
\vec{A_j^F} = \text{Softmax}(\vec{s(F_{j}^{'}, t)}),
\end{align}
and then sum the defect maps corresponding to $ m $ vanilla images embeddings to obtain the segmentation result $\vec{A^F}$,
\begin{align}
\vec{A^F = \sum_{j=1}^{m} A_j^F}.
\end{align}

\noindent
\textbf{Segmentation based on $\vec{V}$.}
Similar to the operations performed on $\vec{F}$, for the new image embedding $\vec{V}$, we discard the $CLS$ token to obtain $\vec{V}[1:]$ 
to calculate the defect map.
Research indicates that erroneous bright spots often appear in the same non-defective areas regardless of the textual prompts. Identifying and removing these irrelevant bright spots as redundant features can effectively reduce noise in the predicted segmentation results \cite{li2023clip}.
Taking the output of the $j$-th encoding block $\vec{V_{j}}$ as an example, the specific operations are as follows:

First, perform $L2$ normalization on the image embedding $\vec{V}[1:]$ and text embedding $\vec{t}$, and then conduct element-wise multiplication to generate a multiplied feature $\vec{V_{j}^{m}}$ containing information from both image and text:
\begin{align}
\vec{V_j^m} = \frac{\vec{V_j}}{\|\vec{V_j\|_2}} \odot \frac{\vec{t}}{\|\vec{t\|_2}}.
\end{align}

We calculate the mean of the multiplied feature $\vec{V_{j}^{m}}$ to obtain the redundant feature $\vec{V_j^r}$ :
\begin{align}
\vec{V}_j^r = \text{mean}(\vec{V}_j^m),
\end{align}
then remove the redundant feature $\vec{V_j^r}$ from the multiplied feature $\vec{V_{j}^{m}}$ to get the defect map:
\begin{align}
\vec{A^V_j} = \text{Softmax}(\vec{V_j^m} - \vec{V_j^r}). 
\end{align}

Sum defect maps corresponding to $m$ new image embeddings $\vec{V}$ to get the segmentation result $\vec{A^V}$:
\begin{align}
\vec{A^V} = \vec{\sum_{j=1}^{m} A^V_j}.
\end{align}

Considering the segmentation results from these two image embeddings, the final overall defect map is given by:
\begin{align}
\vec{A} = \vec{A^F} + \vec{A^V}.
\end{align}

\subsection{Defect Classification}
The self-supervised contrastive learning ability of CLIP \cite{radford2021learning} enables it to understand the semantic relationships between images and text, thereby possessing zero-shot classification capability.
Specifically, the CLIP model encodes the query image $\vec{X}$ to obtain image embeddings, then computes the inner product between the image embeddings with all possible text embeddings, obtaining the label corresponding to the maximum inner product as the classification result. Thereby, we can directly utilize \Cref{eq:similarity}. Since there are $m$ different similarity scores corresponding to $m$ different level image embeddings, we take the maximum score as follows:
\begin{align}
\vec{s_{max}} = \text{Max}(\vec{s(F_{j}^{'}, t)}),  j=1, \cdots, m.
\end{align}
The classification prediction probability obtained through similarity calculation is given by:
\begin{align}
\vec{P_{S}}= \text{Softmax}(\vec{s_{max}}).
\end{align}

Although CLIP's contrastive learning capability enables direct completion of image classification tasks, as we mentioned in \Cref{sub:Defect Segmentation}, it is challenging for pre-trained vision-language models to achieve satisfactory performance directly in specific scenarios. Therefore, we require a few SEM defect images for fine-tuning.

Inspired by the Vision Transformer \cite{dosovitskiy2020image}, which utilizes an extra learnable $[CLS]$ embedding to aggregate information from other tokens during the subsequent image encoding process, resulting in a $CLS$ token aggregating global features, we naturally consider using it to implement classification functionality. The $CLS$ token occupies the first encoding position in the vanilla image embedding $\vec{F}$. Since there are $m$ encoding blocks, we obtain m vanilla image embeddings $\vec{F}$. The classification $CLS$ vectors are represented as:
\begin{align}
\vec{F_C} = [\vec{f^{CLS}_1}, \vec{f^{CLS}_2}, \cdots, \vec{f^{CLS}_m}],
\end{align}
After obtaining effective feature vectors $\vec{F_C}$, we then use it to fine-tune a simple classification head, such as a linear classifier, resulting in the classification prediction probability $\vec{P_{C}}$:
\begin{align}
\vec{F_{C}^{'}} = \vec{W \cdot F_C} + \vec{b},\\
\vec{P_{C}} = \text{Softmax}(\vec{F_{C}^{'}}),
\end{align}
here $\vec{W}$ denotes the weight matrix, and $\vec{b}$ signifies the bias of the classification head.

The final classification prediction probabilities are derived from the image-text contrast score calculated by CLIP and the prediction scores of the classification head, expressed as follows:
\begin{align}
\vec{P} =\vec{(1-\alpha)} \cdot \vec{P_{S}} + \vec{\alpha} \cdot \vec{P_{C}},
\label{eq:classification}
\end{align}
where $\vec{\alpha}$ is a scalar weight that balances these two probabilities.

\section{Experiments}
\label{sec:experiments}


\subsection{Experiments Settings}
\label{sec:experiments-settings}



Evaluation metrics include iAUROC, pAUROC, and pixel-level $F_{1}$-$max$ for segmentation, and Accuracy, Precision, Recall, and $F_{1}$ score for classification.
We utilize the LAION-400M-based CLIP model equipped with ViT-B/16+ for our experiments.  The image encoder backbone consists of 12 layers, we divide them into 4 encoding blocks, i.e., m = 4. Thus, each encoding block contains 3 layers, corresponding to 3 dual path blocks, namely, n = 3.
All experiments are conducted on NVIDIA RTX $4090$. 
For fine-tuning strategies, we employ the Adam optimizer for parameter updates. The hyperparameter $\alpha$ in \Cref{eq:classification} is set to $0.8$.

\subsection{Benchmarks and Baselines}
For defect segmentation performance, we primarily compare our method with WinCLIP$+$ \cite{Jeong_2023_CVPR}, PromptAD \cite{li2024promptad}, DRA \cite{ding2022catching}, and AnomalyGPT \cite{gu2024anomalygpt} under a series of few-shot settings. These methods represent popular anomaly detection (AD) approaches and recent \textit{state-of-the-art} (SOTA) AD models. Both WinCLIP and PromptAD are based on CLIP for anomaly detection. Thus, we configure them with ViT-B/16+, pre-trained on LAION-400M. These baselines are introduced in detail in \Cref{sec:preliminaries}. 

\begin{figure*}[tb!]
  \centering
  \includegraphics[width=0.98\textwidth, trim={0cm 5cm 0cm 3cm}, clip]{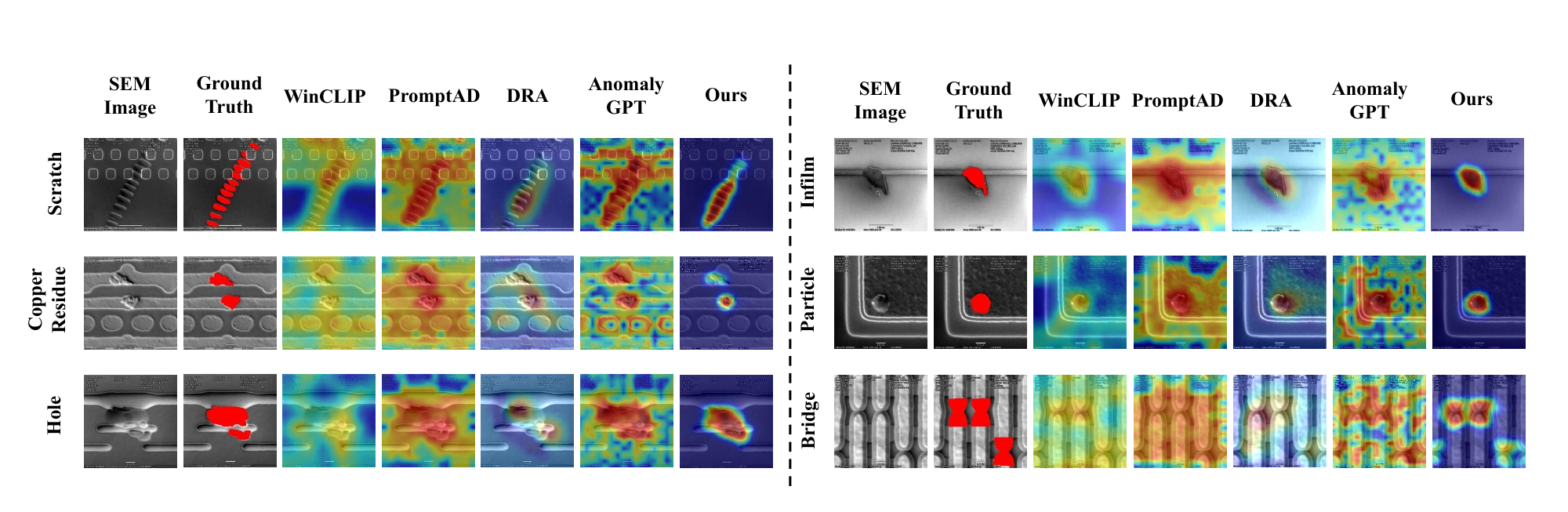}
  \caption{Visualization of 10-shot segmentation.}
  \label{Visualization of 10-shot segmentation}
\end{figure*}

Given the lack of multi-category classification in previous methods, we compare classification performance using models pre-trained on ImageNet-21K \cite{Dataset-2021NeurIPS-ImageNet-21K}, including ViT \cite{dosovitskiy2020image}, ResNet50+ViT \cite{dosovitskiy2020image}, ResNet101 \cite{he2016deep}, and EfficientNet \cite{tan2019efficientnet}. Each model is fine-tuned on our SEM dataset with $10$-shot samples and compared to our SEM-CLIP model on the same test set.

\subsection{Results Analysis}
\noindent
\textbf{Segmentation performance comparisons.}
We evaluated iAUROC, pAUROC, and $F1$-$max$ scores across various shot settings, as shown in \Cref{few-shot-iAUROC}.
The results show that SEM-CLIP outperforms the SOTA scores in BSL across all few-shot settings. Specifically, our method improved by $1.4\uparrow/0.4\uparrow/1.7\uparrow$ in the $1$-shot setting, $1.5\uparrow/0.2\uparrow/11.3\uparrow$ in the $2$-shot setting, $2.1\uparrow/1.9\uparrow/0.4\uparrow$ in the $5$-shot setting, and $1.3\uparrow/0.4\uparrow/1.5\uparrow$ in the $10$-shot setting.

Additionally, under the $10$-shot setting, SEM-CLIP demonstrated precise defect localization and segmentation, effectively distinguishing between normal and defective areas, as shown in \Cref{Visualization of 10-shot segmentation}.

\begin{table}[tb!]
    \caption{Comparison of evaluation metrics (iAUROC/pAUROC/F1-max) under different shot settings (\%).}
    \resizebox{1\linewidth}{!}
    {
        \begin{threeparttable}
            {
                \begin{tabular}{l c c c c}
                    \toprule
                    Models & 1-shot & 2-shot & 5-shot & 10-shot \\
                    \midrule
                     WinCLIP$+$ \cite{Jeong_2023_CVPR} & 51.4/84.5/28.5 & 55.5/85.3/29.5 & 64.9/86.1/29.7 & 72.2/87.0/31.1 \\
                     PromptAD \cite{li2024promptad}           & 94.1/95.8/58.2 & 96.1/96.5/60.4 & 96.3/96.9/61.5 & 97.8/97.3/62.7\\
                     DRA \cite{ding2022catching} & \underline{96.6}/81.2/\underline{67.9} & \underline{97.3}/91.7/\underline{70.5} & \underline{97.6}/\underline{96.9}/\underline{78.2} & \underline{98.5}/\underline{98.2}/\underline{82.3}\\ 
                     AnomalyGPT \cite{gu2024anomalygpt} & 86.8/\underline{96.3}/61.6 & 89.8/\underline{96.6}/63.1 & 86.3/96.5/65.8 &86.4/96.5/65.2 \\
                    \midrule
                   \textbf{SEM-CLIP (Ours)}& \textbf{98.0/96.7/69.6} & \textbf{98.8/96.8/74.4} & \textbf{99.7/97.8/78.6} & \textbf{99.8/98.6/83.8} \\
                    \bottomrule
                \end{tabular}
            }
        \end{threeparttable}
    }
    \label{few-shot-iAUROC}
\end{table}

\noindent
\textbf{Classification performance comparisons.}

\begin{figure}[h]
    \centering
    \includegraphics[width=0.82\linewidth]{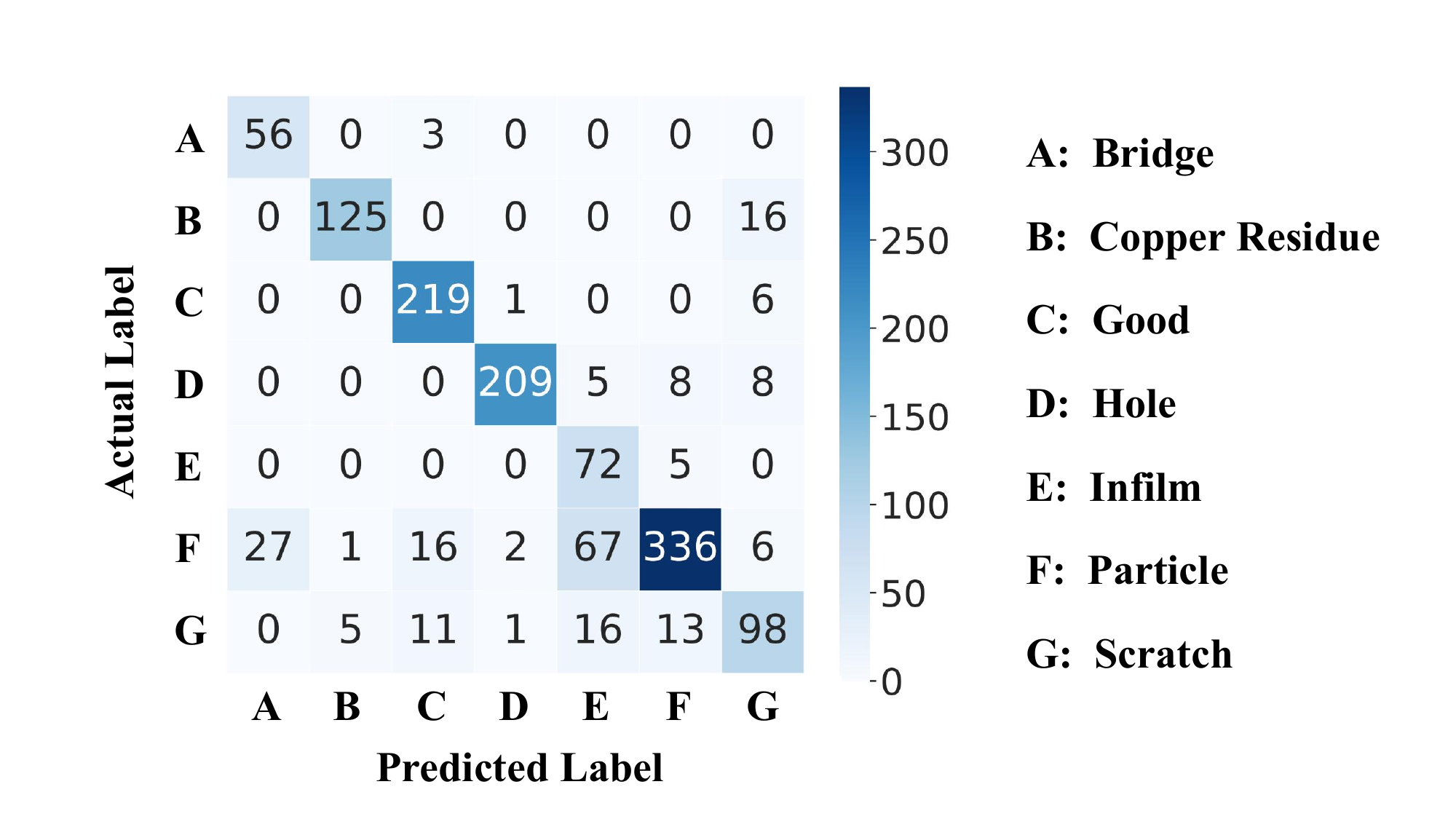}
    \caption{Classification confusion matrix of 10-shot.}
    \label{10-shot matrix}
\end{figure}

SEM-CLIP excels in nearly all metrics, especially in the $F_{1}$ score, demonstrating its ability to identify defect categories while minimizing the false negatives. This makes it ideal for our SEM image classification task involving imbalanced defect categories.
As shown in \Cref{few-shot-class}, our method achieves the highest accuracy, recall, and $F_{1}$ score, although the pre-trained EfficientNet model surpasses ours in precision. This advantage is likely due to EfficientNet's extensive prior knowledge of the diverse ImageNet dataset and advanced regularization techniques. However, EfficientNet's lower overall accuracy suggests weaker recognition capabilities. SEM-CLIP excels in nearly all metrics, particularly in the $F_{1}$ score, highlighting its ability to accurately identify defect categories while minimizing false negatives, making it ideal for SEM image classification with imbalanced categories. The confusion matrix in \Cref{10-shot matrix} shows that SEM-CLIP classifies most defects with high accuracy, though it struggles with the ``particle'' category. This challenge arises from the varied morphologies of particles, which are easily confused with other defects, especially inflim, as these are essentially particles embedded within the film, sharing similar morphology, as shown in \Cref{fig:defect-complexity}.

\begin{table}[tb!]
    \centering
    \caption{Comparison of defect classification performance (\%).}
    \resizebox{0.82\linewidth}{!}
    {
        \begin{threeparttable}
            {
                \begin{tabular}{l|c|c|c|c}
                    \toprule
                    Models & Accuracy & Precision & Recall  & $F_{1}$ \\
                    \midrule
                    ViT \cite{dosovitskiy2020image} & 81.2  & 78.5 & 84.5 & 78.9 \\
                    ResNet101 \cite{he2016deep}  & 71.4  & 72.8 & 76.1 & 70.2 \\
                    ResNet50+ViT \cite{dosovitskiy2020image} & 81.2 & 75.8 & 85.3 & 78.4 \\
                    EfficientNet \cite{tan2019efficientnet}  & 78.5  & \textbf{89.5} & 83.3 & 81.6 \\
                    \midrule
                   \textbf{SEM-CLIP (Ours)} & \textbf{83.7} & 87.2 & \textbf{86.7} & \textbf{84.4} \\
                    \bottomrule
                \end{tabular}
            }
        \end{threeparttable}
    }
    \label{few-shot-class}
\end{table}



\subsection{Abalation Studies}

\begin{figure}[tb!]
    \centering
    \includegraphics[width=0.84\linewidth]{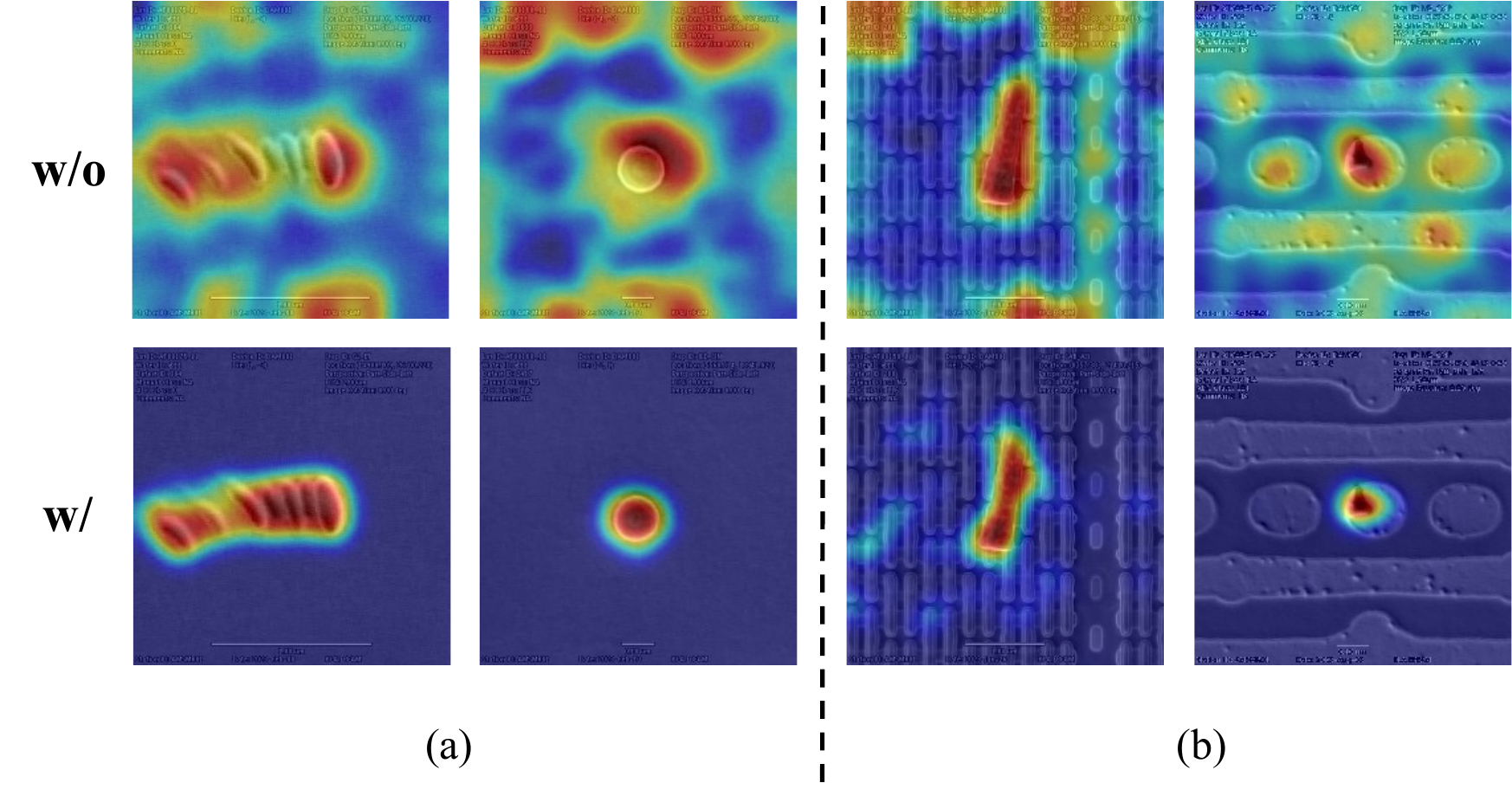}
ok    \caption{Segmentation results w/o (top row) and w/ (bottom row) the Transformation Layer after 10-shot fine-tuning: (a) textual information interference; (b) background patterns interference.}
    \label{fig:interference}
\end{figure}

\noindent
\textbf{SEM-CLIP for defect Segmentation.}
We first examined the impact of fine-tuning with few-shot samples. In \Cref{Ablation}, ``w/o Transformation Layer'' indicates that the Transformation Layer was not used, resulting in direct use of \( \vec{F_{j}} \) for segmentation, as shown in \Cref{fig:interference}. 
Our SEM images are captured from the production line and display textual information at the top and bottom of the image. Without fine-tuning, the model tends to identify this textual information as defects erroneously. Furthermore, the lack of understanding regarding the complexity of SEM image backgrounds also makes it susceptible to mistakenly classifying normal background patterns as defects.


We also assessed the influence of prompt design. ``w/o Detailed Prompt'' refers to using generic prompts instead of detailed, expert-informed ones. The results show that detailed prompts, like ``\{ \} image with a linear scratch'' are more effective.

 Lastly, we analyzed the role of multi-layer features. Our SEM-CLIP model uses outputs from four encoding blocks, including vanilla and new image embeddings, to compute defect maps. "w/o multi-layer" refers to using only the last encoding block’s outputs. Incorporating multi-layer information significantly improves segmentation performance.

\noindent
\textbf{SEM-CLIP for defect Classification.}
\Cref{Ablation} shows the effects of various components on classification. ``w/o \( \vec{P_S} \)'' indicates the exclusion of CLIP’s prior knowledge, leading to classification based solely on the classification head, as in \Cref{eq:classification} with \( \alpha = 1 \). ``w/o \( \vec{P_C} \)'' relies only on text prompt-guided predictions (\( \alpha = 0 \)). The results demonstrate that solely relying on pre-trained CLIP is inadequate for SEM defect classification. Fine-tuning with few-shot samples significantly improves performance, highlighting the importance of few-shot learning in specialized tasks.
For classification, ``w/o multi-layer'' refers to using only the last layer’s CLS token. The results show that employing a multi-layer approach enhances feature detection, leading to superior classification performance by capturing both global and local image features.

\begin{table}[tb!]
    \centering
    \caption{Ablation Studies under the 10-Shot setting.}
    \resizebox{\linewidth}{!}
    {
        \begin{threeparttable}
            {
                \begin{tabular}{c c c c c c c c}
                    \toprule
                    \multirow{2}{*}{Methods} & \multicolumn{3}{c}{Segmentation (\%)} & \multicolumn{4}{c}{Classification (\%)} \\
                    \cmidrule(r){2-4} \cmidrule(l){5-8}
                    & iAUROC & pAUROC & $F_{1}$-$max$ & Acc. & Prec. & Recall & $F_{1}$ \\
                    \midrule
                    w/o Transformation Layer & 86.8 & 79.2 & 29.6 & - & - & - & - \\
                    w/o Detailed Prompt & 99.6 & 98.1 & 82.1 & - & - & - & -  \\
                    w/o $\vec{P_{S}}$  & - & - & - & 83.6 & 87.2 & 86.6 & 84.3 \\
                    w/o $\vec{P_{C}}$  & - & - & - & 25.7 & 20.2 & 30.1 & 16.0 \\
                    w/o multi-layer & 99.4 & 96.2 & 75.9 & 80.5 & 75.6 & 83.8 & 77.7 \\
                    \midrule
                    \textbf{SEM-CLIP} & \textbf{99.8} & \textbf{98.6} & \textbf{83.8} & \textbf{83.7} & \textbf{87.2} & \textbf{86.7} & \textbf{84.4} \\
                    \bottomrule
                \end{tabular}
            }
        \end{threeparttable}
    }
    \label{Ablation}
\end{table}

\section{Conclusions}
\label{sec:conclusions}


In this paper, we introduce SEM-CLIP, a novel few-shot learning approach that innovatively integrates defect classification and segmentation functionalities.
This method utilizes carefully crafted prompts to optimize the vision-language model for more effective text-guided learning.
Additionally, it features a customized architecture for the distinct needs of segmentation and classification tasks. SEM-CLIP effectively minimizes the impact of complex backgrounds inherent in SEM defect data and addresses the challenges of intricate defect textures.

\section*{Acknowledgments}
This work is supported by the Zhejiang University Education Foundation Qizhen Scholar Foundation, the Zhejiang Provincial Key R\&D Programs (Grant No.~2024C01002, No.~2024SJCZX0031).

\clearpage
{

}


\begin{thebibliography}{10}
\providecommand{\url}[1]{#1}
\csname url@samestyle\endcsname
\providecommand{\newblock}{\relax}
\providecommand{\bibinfo}[2]{#2}
\providecommand{\BIBentrySTDinterwordspacing}{\spaceskip=0pt\relax}
\providecommand{\BIBentryALTinterwordstretchfactor}{4}
\providecommand{\BIBentryALTinterwordspacing}{\spaceskip=\fontdimen2\font plus
\BIBentryALTinterwordstretchfactor\fontdimen3\font minus \fontdimen4\font\relax}
\providecommand{\BIBforeignlanguage}[2]{{%
\expandafter\ifx\csname l@#1\endcsname\relax
\typeout{** WARNING: IEEEtran.bst: No hyphenation pattern has been}%
\typeout{** loaded for the language `#1'. Using the pattern for}%
\typeout{** the default language instead.}%
\else
\language=\csname l@#1\endcsname
\fi
#2}}
\providecommand{\BIBdecl}{\relax}
\BIBdecl

\bibitem{chen2008defect}
S.~Chen, T.~Hu, G.~Liu, Z.~Pu, M.~Li, and L.~Du, ``Defect classification algorithm for ic photomask based on pca and svm,'' in \emph{2008 Congress on Image and Signal Processing}, vol.~1.\hskip 1em plus 0.5em minus 0.4em\relax IEEE, 2008, pp. 491--496.

\bibitem{chang2013hybrid}
C.-F. Chang, J.-L. Wu, and Y.-C. Wang, ``A hybrid defect detection method for wafer level chip scale package images,'' \emph{International Journal on Computer, Consumer and Control}, vol.~2, no.~2, pp. 25--36, 2013.

\bibitem{cheon2019convolutional}
S.~Cheon, H.~Lee, C.~O. Kim, and S.~H. Lee, ``Convolutional neural network for wafer surface defect classification and the detection of unknown defect class,'' \emph{IEEE Transactions on Semiconductor Manufacturing}, vol.~32, no.~2, pp. 163--170, 2019.

\bibitem{ronneberger2015u}
O.~Ronneberger, P.~Fischer, and T.~Brox, ``U-net: Convolutional networks for biomedical image segmentation,'' in \emph{Medical image computing and computer-assisted intervention--MICCAI 2015: 18th international conference, Munich, Germany, October 5-9, 2015, proceedings, part III 18}.\hskip 1em plus 0.5em minus 0.4em\relax Springer, 2015, pp. 234--241.

\bibitem{badrinarayanan2017segnet}
V.~Badrinarayanan, A.~Kendall, and R.~Cipolla, ``Segnet: A deep convolutional encoder-decoder architecture for image segmentation,'' \emph{IEEE transactions on pattern analysis and machine intelligence}, vol.~39, no.~12, pp. 2481--2495, 2017.

\bibitem{nag2022wafersegclassnet}
S.~Nag, D.~Makwana, S.~Mittal, C.~K. Mohan \emph{et~al.}, ``Wafersegclassnet-a light-weight network for classification and segmentation of semiconductor wafer defects,'' \emph{Computers in Industry}, vol. 142, p. 103720, 2022.

\bibitem{zavrtanik2021draem}
V.~Zavrtanik, M.~Kristan, and D.~Sko{\v{c}}aj, ``Draem-a discriminatively trained reconstruction embedding for surface anomaly detection,'' in \emph{Proceedings of the IEEE/CVF International Conference on Computer Vision}, 2021, pp. 8330--8339.

\bibitem{jiang2022masked}
J.~Jiang, J.~Zhu, M.~Bilal, Y.~Cui, N.~Kumar, R.~Dou, F.~Su, and X.~Xu, ``Masked swin transformer unet for industrial anomaly detection,'' \emph{IEEE Transactions on Industrial Informatics}, vol.~19, no.~2, pp. 2200--2209, 2022.

\bibitem{jiang2022softpatch}
X.~Jiang, J.~Liu, J.~Wang, Q.~Nie, K.~Wu, Y.~Liu, C.~Wang, and F.~Zheng, ``Softpatch: Unsupervised anomaly detection with noisy data,'' \emph{Advances in Neural Information Processing Systems}, vol.~35, pp. 15\,433--15\,445, 2022.

\bibitem{wang2023multimodal}
Y.~Wang, J.~Peng, J.~Zhang, R.~Yi, Y.~Wang, and C.~Wang, ``Multimodal industrial anomaly detection via hybrid fusion,'' in \emph{Proceedings of the IEEE/CVF Conference on Computer Vision and Pattern Recognition}, 2023, pp. 8032--8041.

\bibitem{radford2021learning}
A.~Radford, J.~W. Kim, C.~Hallacy, A.~Ramesh, G.~Goh, S.~Agarwal, G.~Sastry, A.~Askell, P.~Mishkin, J.~Clark \emph{et~al.}, ``Learning transferable visual models from natural language supervision,'' in \emph{International conference on machine learning}.\hskip 1em plus 0.5em minus 0.4em\relax PMLR, 2021, pp. 8748--8763.

\bibitem{kirillov2023segany}
A.~Kirillov, E.~Mintun, N.~Ravi, H.~Mao, C.~Rolland, L.~Gustafson, T.~Xiao, S.~Whitehead, A.~C. Berg, W.-Y. Lo, P.~Doll{\'a}r, and R.~Girshick, ``Segment anything,'' \emph{arXiv:2304.02643}, 2023.

\bibitem{Li_Li_Li_Niebles_Hoi_2022}
\BIBentryALTinterwordspacing
D.~Li, J.~Li, H.~Li, J.~C. Niebles, and S.~C. Hoi, ``\BIBforeignlanguage{en-US}{Align and prompt: Video-and-language pre-training with entity prompts},'' in \emph{\BIBforeignlanguage{en-US}{2022 IEEE/CVF Conference on Computer Vision and Pattern Recognition (CVPR)}}, Jun 2022. [Online]. Available: \url{http://dx.doi.org/10.1109/cvpr52688.2022.00490}
\BIBentrySTDinterwordspacing

\bibitem{DNN-2017NeurIPS-Attention}
A.~Vaswani, N.~Shazeer, N.~Parmar, J.~Uszkoreit, L.~Jones, A.~N. Gomez, {\L}.~Kaiser, and I.~Polosukhin, ``Attention is all you need,'' \emph{Advances in neural information processing systems}, vol.~30, 2017.

\bibitem{luddecke2022image}
T.~L{\"u}ddecke and A.~Ecker, ``Image segmentation using text and image prompts,'' in \emph{Proceedings of the IEEE/CVF conference on computer vision and pattern recognition}, 2022, pp. 7086--7096.

\bibitem{zhou2022maskclip}
C.~Zhou, C.~C. Loy, and B.~Dai, ``Extract free dense labels from clip,'' in \emph{European Conference on Computer Vision (ECCV)}, 2022.

\bibitem{Zhou_Zhang_Lei_Liu_Liu_2022}
Z.~Zhou, B.~Zhang, Y.~Lei, L.~Liu, and Y.~Liu, ``\BIBforeignlanguage{en-US}{Zegclip: Towards adapting clip for zero-shot semantic segmentation},'' \emph{\BIBforeignlanguage{en-US}{Cornell University - arXiv,Cornell University - arXiv}}, Dec 2022.

\bibitem{Chen_Si_Zhang_Wang_Wang_Tan_2023}
W.~Chen, C.~Si, Z.~Zhang, L.~Wang, Z.~Wang, and T.~Tan, ``\BIBforeignlanguage{en-US}{Semantic prompt for few-shot image recognition},'' Mar 2023.

\bibitem{Wafer-2020TSM-DCNN}
M.~Saqlain, Q.~Abbas, and J.~Y. Lee, ``A deep convolutional neural network for wafer defect identification on an imbalanced dataset in semiconductor manufacturing processes,'' \emph{IEEE Transactions on Semiconductor Manufacturing}, vol.~33, no.~3, pp. 436--444, 2020.

\bibitem{Wafer-2022TSM-FusionTransformer}
Y.~Wei and H.~Wang, ``Mixed-type wafer defect recognition with multi-scale information fusion transformer,'' \emph{IEEE Transactions on Semiconductor Manufacturing}, vol.~35, no.~2, pp. 341--352, 2022.

\bibitem{Wafer-2023ASPDAC-Survey}
H.~Geng, Q.~Sun, T.~Chen, Q.~Xu, T.-Y. Ho, and B.~Yu, ``Mixed-type wafer failure pattern recognition (invited paper),'' in \emph{2023 28th Asia and South Pacific Design Automation Conference (ASP-DAC)}, 2023, pp. 727--732.

\bibitem{ma2023review}
J.~Ma, T.~Zhang, C.~Yang, Y.~Cao, L.~Xie, H.~Tian, and X.~Li, ``Review of wafer surface defect detection methods,'' \emph{Electronics}, vol.~12, no.~8, p. 1787, 2023.

\bibitem{gao2022review}
Y.~Gao, X.~Li, X.~V. Wang, L.~Wang, and L.~Gao, ``A review on recent advances in vision-based defect recognition towards industrial intelligence,'' \emph{Journal of Manufacturing Systems}, vol.~62, pp. 753--766, 2022.

\bibitem{han2020polycrystalline}
H.~Han, C.~Gao, Y.~Zhao, S.~Liao, L.~Tang, and X.~Li, ``Polycrystalline silicon wafer defect segmentation based on deep convolutional neural networks,'' \emph{Pattern Recognition Letters}, vol. 130, pp. 234--241, 2020.

\bibitem{de2023semi}
V.~De~Ridder, B.~Dey, S.~Halder, and B.~Van~Waeyenberge, ``Semi-diffusioninst: A diffusion model based approach for semiconductor defect classification and segmentation,'' in \emph{2023 International Symposium ELMAR}.\hskip 1em plus 0.5em minus 0.4em\relax IEEE, 2023, pp. 61--66.

\bibitem{ding2022catching}
C.~Ding, G.~Pang, and C.~Shen, ``Catching both gray and black swans: Open-set supervised anomaly detection,'' in \emph{Proceedings of the IEEE/CVF Conference on Computer Vision and Pattern Recognition}, 2022.

\bibitem{Jeong_2023_CVPR}
J.~Jeong, Y.~Zou, T.~Kim, D.~Zhang, A.~Ravichandran, and O.~Dabeer, ``Winclip: Zero-/few-shot anomaly classification and segmentation,'' in \emph{Proceedings of the IEEE/CVF Conference on Computer Vision and Pattern Recognition (CVPR)}, June 2023, pp. 19\,606--19\,616.

\bibitem{gu2024anomalygpt}
Z.~Gu, B.~Zhu, G.~Zhu, Y.~Chen, M.~Tang, and J.~Wang, ``Anomalygpt: Detecting industrial anomalies using large vision-language models,'' in \emph{Proceedings of the AAAI Conference on Artificial Intelligence}, vol.~38, no.~3, 2024, pp. 1932--1940.

\bibitem{chen2023clip}
X.~Chen, J.~Zhang, G.~Tian, H.~He, W.~Zhang, Y.~Wang, C.~Wang, Y.~Wu, and Y.~Liu, ``Clip-ad: A language-guided staged dual-path model for zero-shot anomaly detection,'' \emph{arXiv preprint arXiv:2311.00453}, 2023.

\bibitem{li2024promptad}
Y.~Li, A.~Goodge, F.~Liu, and C.-S. Foo, ``Promptad: Zero-shot anomaly detection using text prompts,'' in \emph{Proceedings of the IEEE/CVF Winter Conference on Applications of Computer Vision}, 2024, pp. 1093--1102.

\bibitem{dosovitskiy2020image}
A.~Dosovitskiy, L.~Beyer, A.~Kolesnikov, D.~Weissenborn, X.~Zhai, T.~Unterthiner, M.~Dehghani, M.~Minderer, G.~Heigold, S.~Gelly \emph{et~al.}, ``An image is worth 16x16 words: Transformers for image recognition at scale,'' \emph{arXiv preprint arXiv:2010.11929}, 2020.

\bibitem{li2023clip}
Y.~Li, H.~Wang, Y.~Duan, and X.~Li, ``Clip surgery for better explainability with enhancement in open-vocabulary tasks,'' \emph{arXiv preprint arXiv:2304.05653}, 2023.

\bibitem{Dataset-2021NeurIPS-ImageNet-21K}
T.~Ridnik, E.~Ben-Baruch, A.~Noy, and L.~Zelnik-Manor, ``Imagenet-21k pretraining for the masses,'' \emph{Annual Conference on Neural Information Processing Systems (NeurIPS)}, 2021.

\bibitem{he2016deep}
K.~He, X.~Zhang, S.~Ren, and J.~Sun, ``Deep residual learning for image recognition,'' in \emph{Proceedings of the IEEE conference on computer vision and pattern recognition}, 2016, pp. 770--778.

\bibitem{tan2019efficientnet}
M.~Tan and Q.~Le, ``Efficientnet: Rethinking model scaling for convolutional neural networks,'' in \emph{International conference on machine learning}.\hskip 1em plus 0.5em minus 0.4em\relax PMLR, 2019, pp. 6105--6114.

\end{thebibliography}
\end{document}